\documentclass{article} 
\usepackage{iclr2026_conference,times}

\usepackage{amsmath,amsfonts,bm}

\def\eqref#1{equation~\ref{#1}}

\def\1{\bm{1}}

\DeclareMathAlphabet{\mathsfit}{\encodingdefault}{\sfdefault}{m}{sl}
\SetMathAlphabet{\mathsfit}{bold}{\encodingdefault}{\sfdefault}{bx}{n}

\usepackage{hyperref}
\usepackage{url}
\usepackage[utf8]{inputenc}
\usepackage[T1]{fontenc} 
\usepackage{geometry}
\usepackage{graphicx}
\usepackage{amsmath}
\usepackage{amsfonts}
\usepackage[ruled,vlined,linesnumbered]{algorithm2e}
\usepackage{caption}
\usepackage{color}
\usepackage{threeparttable}
\usepackage{booktabs}
\usepackage{tcolorbox}
\tcbuselibrary{listings, breakable}

\title{LOCA: Logical Chain Augmentation for Scientific Corpus Cleaning}

\author{You-Le Fang$^{1,}$\thanks{These authors contributed equally to this work.} ~ Dong-Shan Jian$^{1,*}$ ~ Xiang Li$^{1,*}$\\ \textbf{Ce Meng}$^{1}$ ~ \textbf{Ling-Shi Meng}$^{1}$ ~ \textbf{Chen-Xu Yan}$^{1}$ ~ \textbf{Zhi-Zhang Bian}$^{1}$ ~ \textbf{Yan-Qing Ma}$^{1,2,}$\thanks{Corresponding author.} \\
$^{1}$School of Physics, Peking University, Beijing 100871, China\\
$^{2}$Center for High Energy Physics, Peking University, Beijing 100871, China\\
\texttt{\{eden,dsjian\}@stu.pku.edu.cn} \\
\texttt{\{lix-PHY,yqma\}@pku.edu.cn}
}

\iclrfinalcopy 

\newtcolorbox{mytextbox}[1][]{
    colback=white,
    colframe=brown,
    fonttitle=\bfseries,
    rounded corners,
    breakable,
    before upper={
        \setlength{\parskip}{1em}
        \setlength{\parindent}{0pt}
    },
    #1 
}

\begin{document}

\maketitle

\begin{abstract}
While Large Language Models (LLMs) excel in general domains, their reliability often falls short in scientific problem-solving. The advancement of scientific AI depends on large-scale, high-quality corpora. However, existing scientific question-answering (QA) datasets suffer from high error rates, frequently resulting from logical leaps and implicit reasoning within the answers. To address this issue, we introduce LOCA (Logical Chain Augmentation), a novel framework for automatically cleaning scientific corpora, implemented through an augment-and-review loop. At its core, LOCA enhances raw answers by completing missing logical steps and explicitly separating the underlying scientific principle from its subsequent derivation. By applying LOCA to challenging scientific corpora, we demonstrate that it can automatically filter noisy datasets, typically reducing the error rate from as high as 20\% to below 2\%. LOCA provides a scalable and effective methodology for creating high-quality scientific corpora, paving the way for more reliable training and evaluation of scientific AI.\footnote{The source code is available at \url{https://github.com/Science-Discovery/LOCA}.}
\end{abstract}

\section{Introduction}
\label{sec:introduction}
LLMs have demonstrated remarkable capabilities across a wide range of general tasks\citep{brown2020language,chatgpt,achiam2023gpt,anil2023palm,touvron2023llama,touvron2023llama2,liu2024deepseek,guo2025deepseek,Claude,comanici2025gemini,gpt5,gpto4,qwq32b,team2025kimi,yang2025qwen3}. However, their reliability often drops in domains demanding extreme accuracy, such as scientific problem-solving. The grand challenge of building artificial general intelligence capable of scientific discovery depends critically on the availability of high-quality, large-scale scientific corpora\citep{zheng2023large}. These corpora, typically structured as QA pairs, are the foundation for training more capable models and for creating reliable benchmarks to evaluate their progress\citep{huang2024large}.

Scientific problems, particularly in fields like physics, present unique difficulties. Unlike well-structured domains like mathematics or code, they require modeling complex real-world scenarios and unstructured reasoning, where the validity of a step depends heavily on context-sensitive scientific principles \citep{qiu2025phybench,feng2025physics,zhang2025abench,siddique2025physicseval,zhang2025physreason}. This complexity leads to the frequent occurrence of errors in existing scientific corpora. For example, our own expert analysis reveals that error rates in major benchmarks' QA pairs can exceed 20\% \citep{qiu2025phybench,feng2025physics}.

We identify a primary reason for this unreliability: the logical incompleteness of provided solutions. Experts and AIs alike tend to omit steps they consider "obvious", inadvertently hiding subtle but critical flaws in their reasoning. While manual expert review offers a gold standard in principle, it is prohibitively slow and expensive, making it unable to scale. While some automated cleaning pipelines exist, they often focus on surface-level answer inconsistency \citep{wang2022self,singh2023beyond,chen2024self,guo2025synthetic,hao2025reformulation,riaz2025metasynth}, failing to address the core issue in scientific reasoning: structural soundness and completeness of the logical argument itself.

To address this issue, we introduce LOCA, a novel framework for scientific corpus cleaning. The key insight is to enforce a detailed, structured and verifiable reasoning process by interpolating missing logical steps and decomposing each step into two orthogonal components: the underlying principle (e.g., a physical law) and its subsequent derivation (e.g., solving an equation). This structured augmentation not only improves the inherent clarity and correctness of the solutions but also enables a highly reliable review loop to flag suspicious QA pairs, and thus resulting in a high-accuracy cleaned corpus. 
Those rejected pairs, now with enhanced structure and readability, become significantly efficient to human expert review and correction.

We evaluate LOCA's performance on challenging physics QA pairs drawn from some existing high-quality corpora. Our experiments show that LOCA significantly reduces the residual error rate of the resulting cleaned corpus (< 2\%), outperforming various baselines.

The contributions of our work can be summarized as follows:
\begin{itemize}
\item We propose LOCA, a novel framework that cleans scientific corpora by enforcing logical completeness and decomposing reasoning steps into verifiable principles and derivations, within an augment-and-review loop.
\item We show that LOCA significantly reduces the residual error rate of the resulting cleaned corpus while retaining a large accepted set, outperforming various baselines.
\item Our work offers a scalable path to building high-quality corpora for reliably training and evaluating scientific AI. Furthermore, LOCA’s structured outputs can speed up expert content creation and serve as effective educational tools.
\end{itemize}

\section{Related Works}
\label{sec:related}

\paragraph{Corpus Cleaning and Augmentation.} What are commonly referred to as corpus cleaning approaches range from heuristic and model-based filtering \citep{soldaini2024dolma, penedo2024fineweb} to synthetic data generation via seed-based synthesis or corpus rephrasing \citep{abdin2024phi, su2024nemotron}. However, more directly related to our work are methods for reviewing and correcting flawed reasoning. These include self-correction pipelines where a model reflects on its own output \citep{madaan2023self,pan2025codecor} and multi-agent debate frameworks where different agents critique solutions \citep{liang2023encouraging,liu2025breaking,du2023improving}. However, these approaches have known limitations: self-correction often misses subtle errors \citep{huang2023large}, and general debate frameworks can lack the domain-specific structure needed for complex scientific validation \citep{liu2025breaking}.

\paragraph{Physics Corpora.} The inherent complexity of physics, characterized by its demand for multi-step reasoning and precise mathematical modeling, has motivated the development of specialized corpora, often released as benchmarks. These range from expert-curated, competition-level problems with detailed solutions in PHYBench \citep{qiu2025phybench} to large-scale, university-level collections such as PHYSICS \citep{feng2025physics}, ABench-Physics \citep{zhang2025abench}, PhysReason \citep{zhang2025physreason} and PhysicsEval \citep{siddique2025physicseval}. These resources, though valuable for presenting challenging problems and striving for optimal answers, are critically limited by reference solutions that often contain logical leaps and errors. This fundamentally reduces their reliability for model training and evaluation, a gap our work aims to address.

\begin{figure}[!h]
\centering
\includegraphics[width=1.0\linewidth]{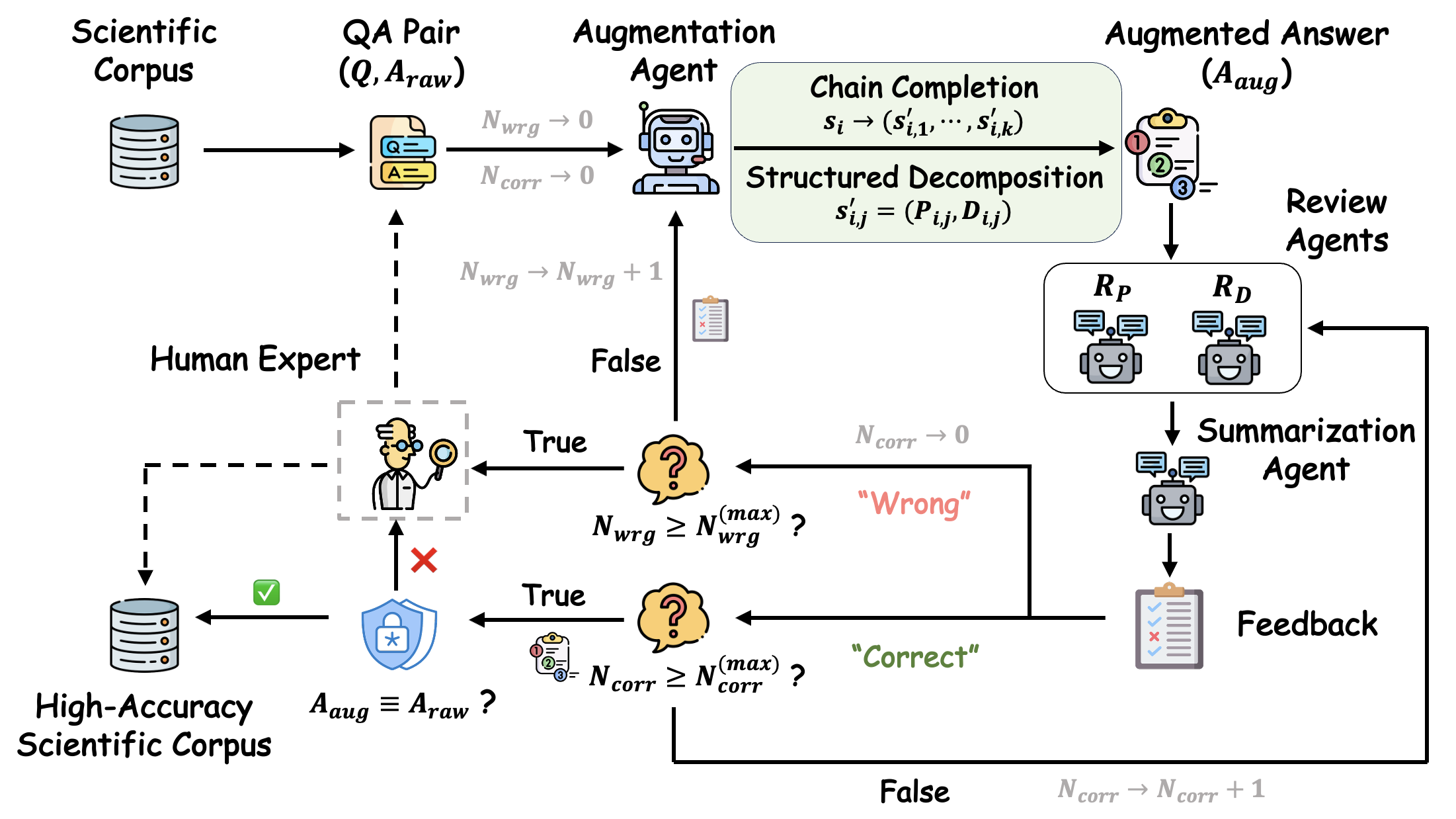}
\caption{\textbf{Pipeline of LOCA.} LOCA employs an iterative augment-and-review loop. In each iteration, given a raw answer with some reasoning process ($A_{\text{aug}}$), an augmentation agent structures it through chain completion and structured decomposition; this structured output is then assessed by specialized review agents. Based on the feedback, the answer is either refined for the next iteration, accepted after passing multiple checks, or rejected. Accepted answers undergo a final external consistency check against $A_{\text{raw}}$, while rejected ones can be flexibly routed to human experts for review.}
\label{fig:framework}
\end{figure}

\section{Method}
\label{sec:method}

We propose LOCA, a framework that cleans and improves scientific corpora by converting the implicit reasoning in each answer into detailed, structured steps that provide a verifiable basis for accepting or rejecting the answer. The pipeline of LOCA is shown in Figure \ref{fig:framework}.

\subsection{Logical Chain Augmentation}

The cornerstone of LOCA is logical chain augmentation, which transforms a raw, unstructured answer with some reasoning process into a structured, verifiable logical chain. Formally, consider a question $Q$ and its raw answer, represented as a sequence of steps $A_{\text{raw}} = (s_1, \dots, s_n)$, where each step $s_i$ represents an implicit transformation from a context $C_{i-1}$ to $C_i$:
\begin{align}
    s_i: C_{i-1} \rightarrow C_i.
\end{align}
Here, $C_n$ contains the final result. Typically, $A_{\text{raw}}$ suffers from two key limitations:
\begin{itemize}
    \item \textbf{Non-Atomicity}. A single step $s_i$ 
  often conflates multiple atomic reasoning steps into a "logical leap," which not only introduces a higher risk of errors but also hinders their precise localization.
    \item \textbf{Implicit Justification}.
     The rationale for a step (the "why," e.g., a physical law) is often entangled with its subsequent mathematical derivation (the "how", e.g., solving an equation) or omitted, obscuring the reasoning process.
\end{itemize}

To address these issues, LOCA employs a transformation implemented through an LLM-based agent, to map the raw answer $A_{\text{raw}}$ to an augmented answer $A_{\text{aug}}$. This transformation is achieved through two intertwined operations: \textbf{chain completion} and \textbf{structured decomposition}.
  \paragraph{Chain Completion.}
This operation enforces atomicity by completing reasoning steps that are often omitted by both human experts and LLMs. 
LOCA addresses this by making all implicit assumptions, intermediate conclusions, and derivations explicit.
  Formally, a step $s_i: C_{i-1} \rightarrow C_i$ is non-atomic if it implicitly contains an intermediate context $C_{\text{int}}$ (i.e., $C_{i-1} \rightarrow C_{\text{int}} \rightarrow C_i$).  LOCA identifies and decomposes each such step into a more fundamental subsequence $(s'_{i,1}, s'_{i,2}, \dots, s'_{i,k})$ where $k > 1$:
  \begin{align}
    \forall s_i \in A_{\text{raw}}, \text{ if } \neg \text{IsAtomic}(s_i), \text{ then } s_i \mapsto (s'_{i,1}, \dots, s'_{i,k}).
  \end{align}
  This results in a new, more detailed sequence $S' = (s'_1, s'_2, \dots, s'_m)$ with $m \ge n$, where each step represents a more atomic inference, significantly reducing the chance of hidden errors within a single step.
  
  \paragraph{Structured Decomposition.} 
  To distinguish the underlying logical assumption from its specific application, we decompose each atomic step $s'_j$ into two orthogonal components: principle ($P_j$) and derivation ($D_j$). This design is inspired by formal proof systems where each inference applies a specific rule to a set of premises \citep{moura2021lean}. Each step in the augmented answer $A_{\text{aug}}$ thus becomes a tuple:
  \begin{align}
    s'_j = (P_j, D_j).
  \end{align}
\begin{itemize}
  \item \textbf{Principle} ($P_j$). A declarative statement of the step's logical foundation 
  from an axiom space $\mathbb{P}$:
  \begin{align}
    P_j \in \mathbb{P} = \{\text{Newton's Second Law, L'Hôpital's Rule, Geometric Constraints ...}\}.
  \end{align}
  $P_j$ answers the question: "Why can this step be taken?"
  \item \textbf{Derivation} ($D_j$). 
  The derivation required to apply principle $P_j$ to the preceding context $C_{j-1}$ and yield the new context $C_j$:
  \begin{equation}
      D_j = \mathcal{D}(C_{j-1}|P_j),\qquad C_j = C_{j-1} \cup P_j \cup D_j.
  \end{equation}
  $D_j$ answers the question: "How is this principle specifically applied?"
\end{itemize}

The final augmented answer is thus a sequence of these structured tuples:
\begin{align}
  A_{\text{aug}} = ((P_1, D_1), (P_2, D_2), \dots, (P_m, D_m)).
\end{align}

\subsection{Iterative Augmentation and Review}
 
The structured feature of $A_{\text{aug}}$ provides a powerful foundation for an iterative cycle of augmentation and review. This loop is designed to progressively improve a solution until it meets a high standard of logical rigor or is definitively flagged as flawed. Notably, the primary goal of this process is not necessarily to increase the answer's correctness, but rather to enable a more accurate \textit{judgment}. Along with the final decision criterion described later, LOCA ensures that the ultimately accepted QA pairs have an exceptionally low residual error rate while maintaining a substantial set size.

In each iteration, the correctness of each step $s'_j$ is reviewed by two specialized components---principle review ($\mathcal{R}_P$) and derivation review ($\mathcal{R}_D$):
\begin{align}
  \mathcal{R}(s'_j) = \mathcal{R}_P(P_j | C_{j-1}) \land \mathcal{R}_D(D_j | P_j, C_{j-1}).
\end{align}
\begin{itemize}
  \item \textbf{Principle Review ($\mathcal{R}_P$)}. This assesses the validity of introducing principle $P_j$ in the given context $C_{j-1}$ (e.g., checking if energy conservation is valid when non-conservative forces exist).
  \item \textbf{Derivation Review ($\mathcal{R}_D$)}. This assesses the correctness of derivation $D_j$, assuming $P_j$ is valid. This is a symbolic or mathematical check ensuring the principle was applied accurately.
\end{itemize}
In practice, we instantiate $\mathcal{R}_P$ and $\mathcal{R}_D$ using specialized LLM agents respectively. This decomposition is crucial as it isolates distinct sources of error, thereby enabling more sensitive error detection. The integrated $\mathcal{R}$ not only yields a binary judgment (correct/wrong) but also provides detailed feedback for further augmentation.

Leveraging this verifiable structure, LOCA employs an iterative augment-and-review loop. The loop terminates once its status is finalized under one of two terminal conditions:
\begin{itemize}
 \item \textbf{Passed}. $A_{\text{aug}}$ passes review when $N_{\text{corr}} \ge N^{(\text{max})}_{\text{corr}}$, where $N_{\text{corr}}$ counts \textit{consecutive} iterations satisfying $\bigwedge_j \mathcal{R}(s'_j)$, and $N^{(\text{max})}_{\text{corr}}$ is a hyperparameter.
 \item \textbf{Failed}. $A_{\text{aug}}$ fails review when $N_{\text{wrg}} \ge N^{(\text{max})}_{\text{wrg}}$, where $N_{\text{wrg}}$ is the \textit{cumulative} count of iterations not satisfying $\bigwedge_j \mathcal{R}(s'_j)$, and $N^{(\text{max})}_{\text{wrg}}$ is a hyperparameter.
\end{itemize}
If $\bigwedge_j \mathcal{R}(s'_j)$ is not satisfied and $N_{\text{wrg}} < N^{(\text{max})}_{\text{wrg}}$ in an iteration, LOCA resets $N_{\text{corr}}=0$ and then leverages the feedback to refine the augmented solution for the next iteration. This mechanism prevents the premature acceptance of flawed answers while using targeted feedback for improvement.
 
\subsection{Final Decision Criterion for Corpus Partitioning}
The LOCA pipeline partitions the initial corpus into two disjoint sets: \textbf{accepted} and \textbf{rejected}. A QA pair is accepted if and only if it satisfies two criteria:
\begin{itemize}
  \item \textbf{Internal Coherence}. The iterative review process must terminate as the ``passed'' state. This confirms the augmented solution is logically stable and self-consistent.
  \item \textbf{External Consistency}. The final result of the augmented answer ($A_{\text{LOCA}}$) must match that of the original answer ($A_{\text{raw}}$), i.e., $A_{\text{LOCA}} \equiv A_{\text{raw}}$. Any mismatch indicates a potential error in either the original or the augmented answer, flagging the QA pair as temporarily unreliable.
\end{itemize}
Notably, the rejected set, containing structured answers $A_{\text{aug}}$, also provides a valuable resource for efficient human expert review and can be added back to the corpus upon correction.

\section{Experiments}
\label{sec:experiments}

To evaluate LOCA's performance, we apply it to a diverse set of physics QA corpora---a representative case in natural sciences---and perform a comprehensive comparison against several mainstream reasoning and review methods. We mainly evaluate each method's capacity to filter a corpus by measuring the residual error rate within their respective accepted sets.

\subsection{Baselines}

The two components of our final decision criterion---external consistency and internal coherence---naturally align with distinct classes of mainstream reasoning and review methods. This correspondence provides a clear basis for structuring our comparison, and we therefore group our baselines into three categories.

\paragraph{Reasoning-Based.}
This category focuses exclusively on the external consistency. For a given problem, these methods generate a new solution from scratch, ignoring the raw answer. A QA pair is accepted only if the newly generated answer, $A_{\text{new}}$, matches the original answer, $A_{\text{raw}}$. 

To establish baselines, we generate $A_{\text{new}}$ by employing several mainstream powerful reasoning strategies:
\begin{itemize}
  \item \textbf{Direct Prompting.} The LLM is prompted to solve the problem in a single pass.
  \item \textbf{Chain-of-Thought (CoT).} We evaluate both Zero-Shot-CoT \citep{kojima2022large}, which appends "Let's think step by step" to the prompt, and Few-Shot CoT \citep{wei2022chain}, which provides in-context examples of step-by-step reasoning.
  \item \textbf{Self-Consistency (CoT-SC).} An extension of CoT where multiple reasoning paths are sampled. The final answer $A_{\text{new}}$ is determined by a majority vote over the outcomes of these paths \citep{wang2022selfconsistency}. We sample $k=5$ paths for this baseline.
  \item \textbf{Tree-of-Thoughts (ToT).} This method moves beyond linear reasoning by exploring a tree of intermediate thoughts. It allows the model to evaluate different reasoning steps \citep{yao2023tree}. We configure ToT with a tree depth of $d=4$ and a node size limit of $k=2$.
  \item \textbf{Graph-of-Thoughts (GoT).} An extension of ToT that models the reasoning process as a more flexible graph structure, allowing for the merging of different reasoning paths \citep{besta2024graph}.
  \item \textbf{Multi-Agent Debate (MAD).} This involves multiple LLM agents that propose and critique solutions in a structured debate format \citep{liang2023encouraging}. We use 2 agents debating for 3 rounds.
\end{itemize}

\paragraph{Review-Based.}
This method isolates the condition of internal coherence in a simplified form. An LLM directly reviews the raw answer to judge its correctness. To mitigate the unreliability of a single review, we adopt a self-consistency approach. A QA pair is accepted only if the LLM judges the answer correct in $N_{\text{corr}}^{(\text{max})}=3$ consecutive, independent evaluations. We denote this method as Review-SC.

\paragraph{Iterative Self-Reflection.}
The third category represents methods that integrate both conditions. We adopt the most representative self-reflection method which employs a feedback-refine loop to improve upon the raw answer \citep{madaan2023self}.  In fact, LOCA itself also belongs to this category and we set $N_{\text{corr}}^{(\text{max})}=3$, $N_{\text{wrg}}^{(\text{max})}=5$ for evaluation. 

\subsection{Datasets}
To evaluate the general applicability of LOCA, we test it on a diverse collection of physics QA pairs sourced from several recent, high-quality physics benchmarks.

\begin{itemize}
\item \textbf{PHYBench \citep{qiu2025phybench}.} This contains competition-level problems created and reviewed by a large group of competition-trained students, consisting of gold medal–level competitors.   For our experiments, we utilize the publicly available set of \textbf{100} problems that include complete, human-authored detailed solutions.
\item \textbf{PHYSICS \citep{feng2025physics}.} This consists of problems from publicly available PhD-qualifying exams. We randomly sample \textbf{100} problems from its text-only questions, specifically excluding proofs tasks.
\item \textbf{ABench-Physics \citep{zhang2025abench}.} This contains university- and competition-level physics problems. Since the original dataset provides only answers without intermediate reasoning, we randomly sample \textbf{100} questions from this set and prompt GPT-5 to produce the corresponding raw answers with reasoning, thereby creating the complete QA pairs for our evaluation. This simulates a realistic scenario of cleaning corpora with LLM-generated reasoning.
\end{itemize}
LOCA's structured, complete logical-chain outputs enable our human experts to efficiently identify and correct erroneous answers in the datasets. Using this method, we manually detected 20, 22, and 13 incorrect answers in the three benchmarks, respectively, and provided correct answers wherever possible. These results serve as the ground-truth for the subsequent evaluation.

\begin{table*}[t]
    \centering
    \caption{\textbf{Performance comparison on PHYBench}. We report the residual error rate (\%) of the accepted set, with its size (number of QA pairs) in parentheses. An ideal method minimizes the error rate, our primary focus, while maximizing the size of accepted set. Bold indicates the best result for each LLM. LOCA with Gemini 2.5 Pro significantly reduces the error rate while retaining a large accepted set.
    }
    \vspace{-0.2cm}
    \begin{threeparttable} 
    \resizebox{0.8\textwidth}{!}{
    \begin{tabular}{@{}c|ccccc@{}} 
    \specialrule{.16em}{0pt}{.65ex}
         Method  &  & Gemini 2.5 Pro & o3 & DeepSeek-R1 & GPT-5 \\
         \specialrule{.07em}{.4ex}{.65ex}

         Direct Prompting & &6.25\% (48) &10.00\% (30) &7.14\% (42) &12.24\% (49) \\
         Zero-Shot-CoT & &11.76\% (51) &12.50\% (32) &9.30\% (43) &15.91\% (44) \\
         Few-Shot CoT & &7.84\% (51) &13.16\% (38) &\textbf{5.71\%} (35) &10.20\% (49) \\
         CoT-SC & &10.42\% (48) &12.12\% (33) &9.09\% (33) &12.90\% (31) \\
         ToT & &9.43\% (53) &16.00\% (25) &8.57\% (35) &8.16\% (49) \\
         GoT & &7.14\% (42) &14.81\% (27) &9.09\% (22) &13.04\% (23) \\
         MAD & &8.57\% (70) &9.62\% (52) &10.71\% (56) &7.02\% (57) \\
         
         Review-SC & &12.66\% (79) &15.38\% (78) &17.05\% (88) &14.63\% (82) \\
         
         Self-Reflection & &10.39\% (77) &12.50\% (80) &16.46\% (79)  &11.84\% (76) \\
         \specialrule{.07em}{.4ex}{.65ex}
         LOCA (ours) & &\boxed{\textbf{1.69\%}} (59) &\textbf{4.26\%} (47) &10.26\% (39) &\textbf{6.56\%} (61) \\

         \specialrule{.16em}{.4ex}{0pt}
    \end{tabular}%
    }
    \end{threeparttable}
    \label{tab:phybench}
\end{table*}

\subsection{Results}
\paragraph{Preliminary Evaluation on PHYBench.}

In Table \ref{tab:phybench}, a comparison across models reveals that Gemini 2.5 Pro generally provides the most favorable balance between achieving a low residual error rate and retaining a substantial number of QA pairs on various methods used. Besides, the combination of LOCA with Gemini 2.5 Pro achieves the lowest residual error rate among all combinations. Given the superior and robust performance of Gemini 2.5 Pro, we representatively select it for subsequent comprehensive experiments.

\begin{table*}[t]
    \centering
    \caption{\textbf{Comprehensive comparison across various datasets.} Gemini 2.5 Pro is used for all cases, and results are also presented as: residual error rate (\%) (accepted set size). Bold indicates the best method for each dataset. LOCA consistently achieves the lowest error rates across three datasets, demonstrating robust performance.}
    \vspace{-0.2cm}
    \begin{threeparttable} 
    \resizebox{0.65\textwidth}{!}{%
    \begin{tabular}{@{}c|cccc@{}}
    \specialrule{.16em}{0pt}{.65ex}
         Method  &  & PHYBench & PHYSICS & ABench-Physics\\
         \specialrule{.07em}{.4ex}{.65ex}

         Direct Prompting & &6.25\% (48) &9.52\% (63) &10.26\% (78) \\
         Zero-Shot-CoT & &11.76\% (51) &10.17\% (59) &10.00\% (80)  \\
         Few-Shot CoT & &7.84\% (51) &6.67\% (60) &11.69\% (77)  \\
         CoT-SC & &10.42\% (48) &8.20\% (61) &6.67\% (75)  \\
         ToT & &9.43\% (53) &6.90\% (58) &7.50\% (80)  \\
         GoT & &7.14\% (42) &6.78\% (59) &6.67\% (75)  \\
         MAD & &8.57\% (70) &6.25\% (48) &8.70\% (69)  \\
         
         Review-SC & &12.66\% (79) &12.35\% (81)	&8.51\% (94) \\
         
         Self-Reflection & &10.39\% (77) &11.11\% (72)	&7.61\% (92)		  \\
         \specialrule{.07em}{.4ex}{.65ex}
         LOCA (ours) & &\textbf{1.69\%} (59) &\textbf{1.64\%} (61)	&\textbf{1.22\%} (82)		  \\

         \specialrule{.16em}{.4ex}{0pt}
    \end{tabular}%
    }
    \end{threeparttable}
    \label{tab:math_results}
\end{table*}

\paragraph{Comprehensive Evaluation across Various Datasets.}
As shown in Table \ref{tab:math_results}, LOCA substantially outperforms all baselines across three datasets. It significantly reduces the residual error rate to below 2\% while retaining a substantial number of accepted QA pairs. The most telling comparison is against Self-Reflection \citep{madaan2023self}, which shares a similar framework with LOCA but lacks our logical chain augmentation. Self-Reflection yields a $\sim$10 times larger error rate. Besides, the performance of baselines also underscores the necessity of LOCA's final decision criterion. Reasoning-based methods, relying solely on external consistency, exhibit high error rates (ranging from 6.25\% to 11.76\%). This approach can create a ``false agreement'', where $A_{\text{new}} \equiv A_{\text{raw}}$ not because both are correct, but because they might share a common mistake. Similarly, Review-SC, which isolates internal coherence, also proves unreliable (8.51\% to 12.66\%).

\begin{table*}[t]
    \centering
    \caption{\textbf{Ablation study on LOCA's core components.} We evaluate variants by replacing the structured augmentation module, the specialized review module, or both. The results demonstrate that both components are critical and interdependent for minimizing the residual error rate.
    }
    \vspace{-0.2cm}
    \begin{threeparttable} 
    \resizebox{0.75\textwidth}{!}{%
    \begin{tabular}{@{}c|cccc@{}}
    \specialrule{.16em}{0pt}{.65ex}
         Method  &  & PHYBench & PHYSICS & ABench-Physics\\
         \specialrule{.07em}{.4ex}{.65ex}
         w/o Structured Augmentation & &10.14\% (69) &6.35\% (63)	&4.82\% (83)		  \\
         w/o Specialized Review & &4.48\% (67) &9.59\% (73)	&4.71\% (85)		  \\
         w/o Both & &4.62\% (65) &6.15\% (65)	&6.59\% (91)		  \\
         \specialrule{.07em}{.4ex}{.65ex}
         LOCA (ours) & &\textbf{1.69\%} (59) &\textbf{1.64\%} (61)	&\textbf{1.22\%} (82)		  \\

         \specialrule{.16em}{.4ex}{0pt}
    \end{tabular}%
    }
    \end{threeparttable}
    \label{tab:ablation}
\end{table*}

\paragraph{Ablation Study.}
To evaluate the individual contributions of LOCA's two core components---structured augmentation and specialized review---we conduct a comprehensive ablation study. In this study,  we replace the structured augmentation and/or review module(s) with generic counterparts: a feedback-based refine module and a holistic review module. As shown in Table \ref{tab:ablation}, replacing either component leads to a significant performance collapse. For instance, on PHYBench, the error rate increases from 1.69\% to 10.14\% without structured augmentation; on PHYSICS, replacing only the specialized review yields a 9.59\% error rate, which is even worse than replacing both modules. This highlights a crucial point: the components are not merely additive but are tightly coupled. The observation that replacing just one part can be sometimes more harmful than replacing both underscores their strong interdependence.  This validates that the co-design is essential to LOCA's effectiveness.

\begin{figure}[!ht]
\centering
\includegraphics[width=0.7\linewidth]{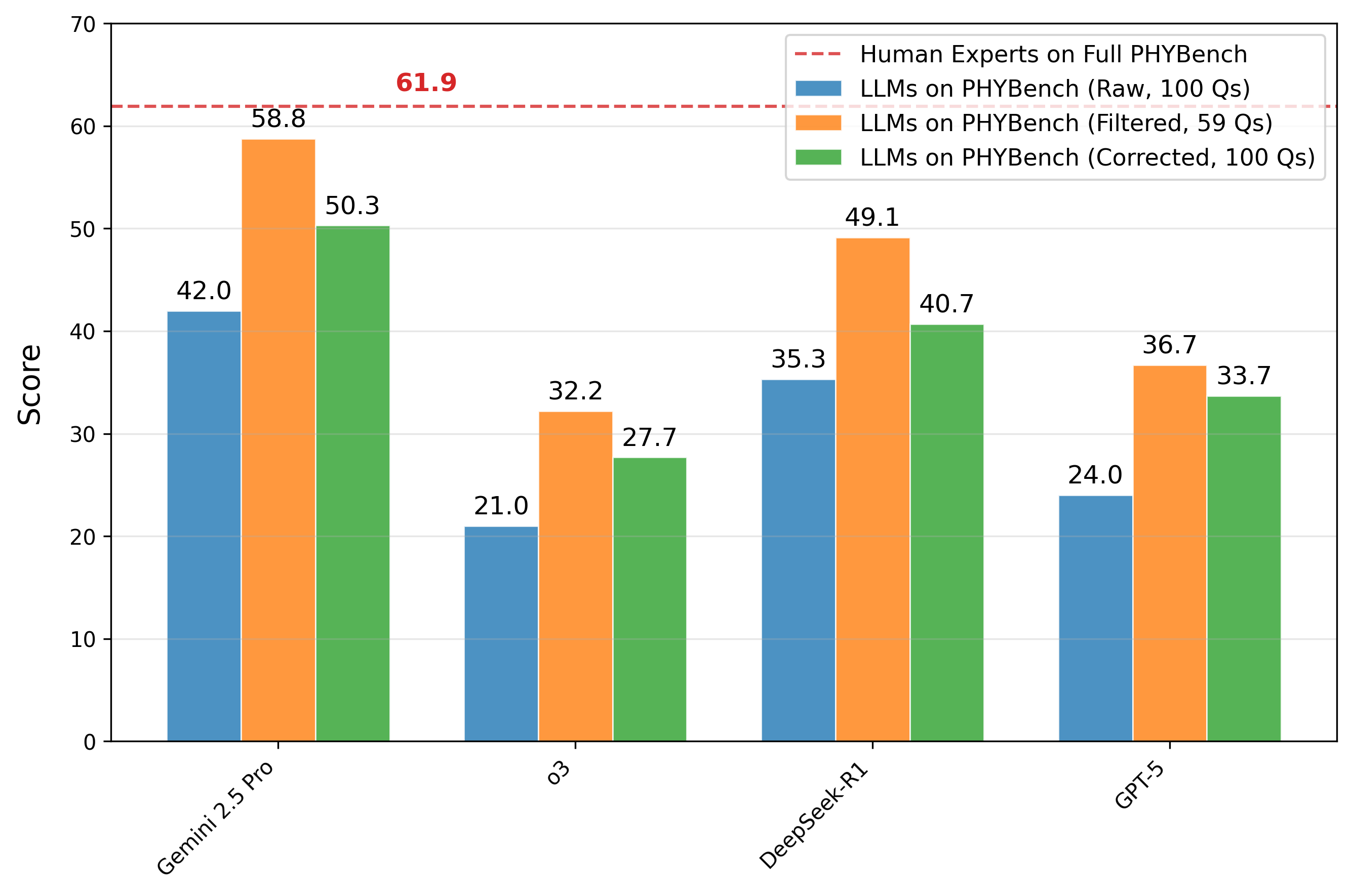}
\caption{\textbf{Impact on evaluating LLMs' performance.} We compare model performance on 3 versions of PHYBench: the original Raw set (100 Qs); the high-accuracy Filtered subset accepted by LOCA (59 Qs); and the Corrected set (100 Qs), where flawed QA pairs are manually fixed powered by LOCA’s augmentation.}
\label{fig:scores_comparison}
\end{figure}

\paragraph{Impact of LOCA's Cleaning on Model Capability Evaluation.}
LOCA enables a more faithful and accurate evaluation of LLM performance. To demonstrate this, we evaluate several models on 3 versions of PHYBench, as shown in Figure \ref{fig:scores_comparison}: (1) the original Raw set; (2) a Filtered high-accuracy subset containing only QA pairs accepted by LOCA with Gemini 2.5 Pro; and (3) a Corrected version where flawed QA pairs have been fixed by human experts, powered by LOCA's augmentation. The results reveal two key findings. First, all models achieve their highest scores on the Filtered subset, which is expected as this set excludes potentially problematic or harder questions. More importantly, model scores on the Corrected benchmark are also substantially higher than on the Raw version. For instance, Gemini 2.5 Pro's score increases from 42.0 to 50.3, closer to the human baseline (61.9) reported \citep{qiu2025phybench}. This score improvement, derived from the same set of 100 questions, confirms that the original benchmark's errors are masking the model's true performance. These corrected scores provide a more holistic and accurate measure of a model's capabilities. Our findings underscore that a robust process like LOCA is critical not just for filtering, but for creating more reliable benchmarks.

\section{Conclusion}
\label{sec:conclusion}

In this work, we introduced LOCA, a novel framework designed to address the issue of high error rates in scientific QA corpora. By automatically enforcing logical completeness and decomposing complex reasoning into explicit principles and derivations, LOCA employs an augment-and-review loop to effectively filter noisy datasets while refining them for straightforward human expert verification. Our comprehensive experiments on challenging physics problems demonstrate that LOCA can reduce the residual error rate from typically 20\% to below 2\%, substantially outperforming a wide range of mainstream reasoning and review baselines. Furthermore, we show that benchmarks cleaned by LOCA enable a more accurate and faithful evaluation of LLM capabilities.  

Although demonstrated in physics, LOCA's methodology is broadly applicable to other principle-based fields like chemistry and engineering. By providing a robust pathway to generating large-scale, high-quality scientific corpora, LOCA represents a key advance toward more trustworthy scientific AI. The structured, logically coherent outputs it produces also hold significant educational value.

\paragraph{Limitations.}
LOCA focuses on checking the correctness of \textit{answers}. A limitation is that a small fraction of questions in scientific corpora can be ill-defined, ambiguous, or factually incorrect. Such flawed questions can interfere with the review process, as the notion of a ``correct'' solution becomes ambiguous.

\subsubsection*{Acknowledgments}

This work is supported by the National Natural Science Foundation of China (No. 12325503), and the High-performance Computing Platform of Peking University.

\bibliography{iclr2026_conference}

\begin{thebibliography}{44}
\providecommand{\natexlab}[1]{#1}
\providecommand{\url}[1]{\texttt{#1}}
\expandafter\ifx\csname urlstyle\endcsname\relax
  \providecommand{\doi}[1]{doi: #1}\else
  \providecommand{\doi}{doi: \begingroup \urlstyle{rm}\Url}\fi

\bibitem[Abdin et~al.(2024)Abdin, Aneja, Behl, Bubeck, Eldan, Gunasekar,
  Harrison, Hewett, Javaheripi, Kauffmann, et~al.]{abdin2024phi}
Marah Abdin, Jyoti Aneja, Harkirat Behl, S{\'e}bastien Bubeck, Ronen Eldan,
  Suriya Gunasekar, Michael Harrison, Russell~J Hewett, Mojan Javaheripi, Piero
  Kauffmann, et~al.
\newblock Phi-4 technical report.
\newblock \emph{arXiv preprint arXiv:2412.08905}, 2024.

\bibitem[Achiam et~al.(2023)Achiam, Adler, Agarwal, Ahmad, Akkaya, Aleman,
  Almeida, Altenschmidt, Altman, Anadkat, et~al.]{achiam2023gpt}
Josh Achiam, Steven Adler, Sandhini Agarwal, Lama Ahmad, Ilge Akkaya,
  Florencia~Leoni Aleman, Diogo Almeida, Janko Altenschmidt, Sam Altman,
  Shyamal Anadkat, et~al.
\newblock Gpt-4 technical report.
\newblock \emph{arXiv preprint arXiv:2303.08774}, 2023.

\bibitem[Anil et~al.(2023)Anil, Dai, Firat, Johnson, Lepikhin, Passos, Shakeri,
  Taropa, Bailey, Chen, et~al.]{anil2023palm}
Rohan Anil, Andrew~M Dai, Orhan Firat, Melvin Johnson, Dmitry Lepikhin,
  Alexandre Passos, Siamak Shakeri, Emanuel Taropa, Paige Bailey, Zhifeng Chen,
  et~al.
\newblock Palm 2 technical report.
\newblock \emph{arXiv preprint arXiv:2305.10403}, 2023.

\bibitem[Besta et~al.(2024)Besta, Blach, Kubicek, Gerstenberger, Podstawski,
  Gianinazzi, Gajda, Lehmann, Niewiadomski, Nyczyk, et~al.]{besta2024graph}
Maciej Besta, Nils Blach, Ales Kubicek, Robert Gerstenberger, Michal
  Podstawski, Lukas Gianinazzi, Joanna Gajda, Tomasz Lehmann, Hubert
  Niewiadomski, Piotr Nyczyk, et~al.
\newblock Graph of thoughts: Solving elaborate problems with large language
  models.
\newblock In \emph{Proceedings of the AAAI conference on artificial
  intelligence}, volume~38, pp.\  17682--17690, 2024.

\bibitem[Brown et~al.(2020)Brown, Mann, Ryder, Subbiah, Kaplan, Dhariwal,
  Neelakantan, Shyam, Sastry, Askell, et~al.]{brown2020language}
Tom Brown, Benjamin Mann, Nick Ryder, Melanie Subbiah, Jared~D Kaplan, Prafulla
  Dhariwal, Arvind Neelakantan, Pranav Shyam, Girish Sastry, Amanda Askell,
  et~al.
\newblock Language models are few-shot learners.
\newblock \emph{Advances in neural information processing systems},
  33:\penalty0 1877--1901, 2020.

\bibitem[Chen et~al.(2024)Chen, Deng, Yuan, Ji, and Gu]{chen2024self}
Zixiang Chen, Yihe Deng, Huizhuo Yuan, Kaixuan Ji, and Quanquan Gu.
\newblock Self-play fine-tuning converts weak language models to strong
  language models.
\newblock \emph{arXiv preprint arXiv:2401.01335}, 2024.

\bibitem[claude(2025)]{Claude}
claude.
\newblock Claude 4.
\newblock \url{https://www.anthropic.com/news/claude-4}, 2025.

\bibitem[Comanici et~al.(2025)Comanici, Bieber, Schaekermann, Pasupat,
  Sachdeva, Dhillon, Blistein, Ram, Zhang, Rosen, et~al.]{comanici2025gemini}
Gheorghe Comanici, Eric Bieber, Mike Schaekermann, Ice Pasupat, Noveen
  Sachdeva, Inderjit Dhillon, Marcel Blistein, Ori Ram, Dan Zhang, Evan Rosen,
  et~al.
\newblock Gemini 2.5: Pushing the frontier with advanced reasoning,
  multimodality, long context, and next generation agentic capabilities.
\newblock \emph{arXiv preprint arXiv:2507.06261}, 2025.

\bibitem[Du et~al.(2023)Du, Li, Torralba, Tenenbaum, and
  Mordatch]{du2023improving}
Yilun Du, Shuang Li, Antonio Torralba, Joshua~B Tenenbaum, and Igor Mordatch.
\newblock Improving factuality and reasoning in language models through
  multiagent debate.
\newblock In \emph{Forty-first International Conference on Machine Learning},
  2023.

\bibitem[Feng et~al.(2025)Feng, Zhao, Liu, Yang, Zhao, Sous, and
  Cohan]{feng2025physics}
Kaiyue Feng, Yilun Zhao, Yixin Liu, Tianyu Yang, Chen Zhao, John Sous, and
  Arman Cohan.
\newblock Physics: Benchmarking foundation models on university-level physics
  problem solving.
\newblock \emph{arXiv preprint arXiv:2503.21821}, 2025.

\bibitem[Guo et~al.(2025{\natexlab{a}})Guo, Yang, Zhang, Song, Zhang, Xu, Zhu,
  Ma, Wang, Bi, et~al.]{guo2025deepseek}
Daya Guo, Dejian Yang, Haowei Zhang, Junxiao Song, Ruoyu Zhang, Runxin Xu,
  Qihao Zhu, Shirong Ma, Peiyi Wang, Xiao Bi, et~al.
\newblock Deepseek-r1: Incentivizing reasoning capability in llms via
  reinforcement learning.
\newblock \emph{arXiv preprint arXiv:2501.12948}, 2025{\natexlab{a}}.

\bibitem[Guo et~al.(2025{\natexlab{b}})Guo, Guo, Huang, Wang, Zhang, Yu, Zhang,
  and Shen]{guo2025synthetic}
Yiduo Guo, Zhen Guo, Chuanwei Huang, Zi-Ang Wang, Zekai Zhang, Haofei Yu,
  Huishuai Zhang, and Yikang Shen.
\newblock Synthetic data rl: Task definition is all you need.
\newblock \emph{arXiv preprint arXiv:2505.17063}, 2025{\natexlab{b}}.

\bibitem[Hao et~al.(2025)Hao, Zhu, Zhang, Shen, and Li]{hao2025reformulation}
Xintong Hao, Ruijie Zhu, Ge~Zhang, Ke~Shen, and Chenggang Li.
\newblock Reformulation for pretraining data augmentation.
\newblock \emph{arXiv preprint arXiv:2502.04235}, 2025.

\bibitem[Huang et~al.(2024)Huang, Yan, Li, and Peng]{huang2024large}
Dawei Huang, Chuan Yan, Qing Li, and Xiaojiang Peng.
\newblock From large language models to large multimodal models: A literature
  review.
\newblock \emph{Applied Sciences}, 14\penalty0 (12):\penalty0 5068, 2024.

\bibitem[Huang et~al.(2023)Huang, Chen, Mishra, Zheng, Yu, Song, and
  Zhou]{huang2023large}
Jie Huang, Xinyun Chen, Swaroop Mishra, Huaixiu~Steven Zheng, Adams~Wei Yu,
  Xinying Song, and Denny Zhou.
\newblock Large language models cannot self-correct reasoning yet.
\newblock \emph{arXiv preprint arXiv:2310.01798}, 2023.

\bibitem[Kojima et~al.(2022)Kojima, Gu, Reid, Matsuo, and
  Iwasawa]{kojima2022large}
Takeshi Kojima, Shixiang~Shane Gu, Machel Reid, Yutaka Matsuo, and Yusuke
  Iwasawa.
\newblock Large language models are zero-shot reasoners.
\newblock \emph{Advances in neural information processing systems},
  35:\penalty0 22199--22213, 2022.

\bibitem[Liang et~al.(2023)Liang, He, Jiao, Wang, Wang, Wang, Yang, Shi, and
  Tu]{liang2023encouraging}
Tian Liang, Zhiwei He, Wenxiang Jiao, Xing Wang, Yan Wang, Rui Wang, Yujiu
  Yang, Shuming Shi, and Zhaopeng Tu.
\newblock Encouraging divergent thinking in large language models through
  multi-agent debate.
\newblock \emph{arXiv preprint arXiv:2305.19118}, 2023.

\bibitem[Liu et~al.(2024)Liu, Feng, Xue, Wang, Wu, Lu, Zhao, Deng, Zhang, Ruan,
  et~al.]{liu2024deepseek}
Aixin Liu, Bei Feng, Bing Xue, Bingxuan Wang, Bochao Wu, Chengda Lu, Chenggang
  Zhao, Chengqi Deng, Chenyu Zhang, Chong Ruan, et~al.
\newblock Deepseek-v3 technical report.
\newblock \emph{arXiv preprint arXiv:2412.19437}, 2024.

\bibitem[Liu et~al.(2025)Liu, Cao, Li, He, and Tan]{liu2025breaking}
Yexiang Liu, Jie Cao, Zekun Li, Ran He, and Tieniu Tan.
\newblock Breaking mental set to improve reasoning through diverse multi-agent
  debate.
\newblock In \emph{The Thirteenth International Conference on Learning
  Representations}, 2025.

\bibitem[Madaan et~al.(2023)Madaan, Tandon, Gupta, Hallinan, Gao, Wiegreffe,
  Alon, Dziri, Prabhumoye, Yang, et~al.]{madaan2023self}
Aman Madaan, Niket Tandon, Prakhar Gupta, Skyler Hallinan, Luyu Gao, Sarah
  Wiegreffe, Uri Alon, Nouha Dziri, Shrimai Prabhumoye, Yiming Yang, et~al.
\newblock Self-refine: Iterative refinement with self-feedback.
\newblock \emph{Advances in Neural Information Processing Systems},
  36:\penalty0 46534--46594, 2023.

\bibitem[Moura \& Ullrich(2021)Moura and Ullrich]{moura2021lean}
Leonardo~de Moura and Sebastian Ullrich.
\newblock The lean 4 theorem prover and programming language.
\newblock In \emph{International Conference on Automated Deduction}, pp.\
  625--635. Springer, 2021.

\bibitem[OpenAI(2022)]{chatgpt}
OpenAI.
\newblock Introducing chatgpt, 2022.
\newblock \url{https://openai.com/blog/chatgpt}.

\bibitem[OpenAI(2025{\natexlab{a}})]{gpt5}
OpenAI.
\newblock Introducing openai gpt-5.
\newblock \url{https://openai.com/index/introducing-gpt-5/},
  2025{\natexlab{a}}.

\bibitem[OpenAI(2025{\natexlab{b}})]{gpto4}
OpenAI.
\newblock Introducing openai o3 and o4-mini.
\newblock \url{https://openai.com/index/introducing-o3-and-o4-mini/},
  2025{\natexlab{b}}.

\bibitem[Pan et~al.(2025)Pan, Zhang, and Liu]{pan2025codecor}
Ruwei Pan, Hongyu Zhang, and Chao Liu.
\newblock Codecor: An llm-based self-reflective multi-agent framework for code
  generation.
\newblock \emph{arXiv preprint arXiv:2501.07811}, 2025.

\bibitem[Penedo et~al.(2024)Penedo, Kydl{\'\i}{\v{c}}ek, Lozhkov, Mitchell,
  Raffel, Von~Werra, Wolf, et~al.]{penedo2024fineweb}
Guilherme Penedo, Hynek Kydl{\'\i}{\v{c}}ek, Anton Lozhkov, Margaret Mitchell,
  Colin~A Raffel, Leandro Von~Werra, Thomas Wolf, et~al.
\newblock The fineweb datasets: Decanting the web for the finest text data at
  scale.
\newblock \emph{Advances in Neural Information Processing Systems},
  37:\penalty0 30811--30849, 2024.

\bibitem[Qiu et~al.(2025)Qiu, Guo, Song, Sun, Cai, Wei, Luo, Yin, Zhang, Hu,
  et~al.]{qiu2025phybench}
Shi Qiu, Shaoyang Guo, Zhuo-Yang Song, Yunbo Sun, Zeyu Cai, Jiashen Wei, Tianyu
  Luo, Yixuan Yin, Haoxu Zhang, Yi~Hu, et~al.
\newblock Phybench: Holistic evaluation of physical perception and reasoning in
  large language models.
\newblock \emph{arXiv preprint arXiv:2504.16074}, 2025.

\bibitem[Riaz et~al.(2025)Riaz, Bhabesh, Arannil, Ballesteros, and
  Horwood]{riaz2025metasynth}
Haris Riaz, Sourav Bhabesh, Vinayak Arannil, Miguel Ballesteros, and Graham
  Horwood.
\newblock Metasynth: Meta-prompting-driven agentic scaffolds for diverse
  synthetic data generation.
\newblock \emph{arXiv preprint arXiv:2504.12563}, 2025.

\bibitem[Siddique et~al.(2025)Siddique, Alam, Rafy, Raiyan, Mahmud, and
  Hasan]{siddique2025physicseval}
Oshayer Siddique, JM~Alam, Md~Jobayer~Rahman Rafy, Syed~Rifat Raiyan, Hasan
  Mahmud, and Md~Kamrul Hasan.
\newblock Physicseval: Inference-time techniques to improve the reasoning
  proficiency of large language models on physics problems.
\newblock \emph{arXiv preprint arXiv:2508.00079}, 2025.

\bibitem[Singh et~al.(2023)Singh, Co-Reyes, Agarwal, Anand, Patil, Garcia, Liu,
  Harrison, Lee, Xu, et~al.]{singh2023beyond}
Avi Singh, John~D Co-Reyes, Rishabh Agarwal, Ankesh Anand, Piyush Patil, Xavier
  Garcia, Peter~J Liu, James Harrison, Jaehoon Lee, Kelvin Xu, et~al.
\newblock Beyond human data: Scaling self-training for problem-solving with
  language models.
\newblock \emph{arXiv preprint arXiv:2312.06585}, 2023.

\bibitem[Soldaini et~al.(2024)Soldaini, Kinney, Bhagia, Schwenk, Atkinson,
  Authur, Bogin, Chandu, Dumas, Elazar, et~al.]{soldaini2024dolma}
Luca Soldaini, Rodney Kinney, Akshita Bhagia, Dustin Schwenk, David Atkinson,
  Russell Authur, Ben Bogin, Khyathi Chandu, Jennifer Dumas, Yanai Elazar,
  et~al.
\newblock Dolma: An open corpus of three trillion tokens for language model
  pretraining research.
\newblock \emph{arXiv preprint arXiv:2402.00159}, 2024.

\bibitem[Su et~al.(2024)Su, Kong, Lin, Jennings, Norick, Kliegl, Patwary,
  Shoeybi, and Catanzaro]{su2024nemotron}
Dan Su, Kezhi Kong, Ying Lin, Joseph Jennings, Brandon Norick, Markus Kliegl,
  Mostofa Patwary, Mohammad Shoeybi, and Bryan Catanzaro.
\newblock Nemotron-cc: Transforming common crawl into a refined long-horizon
  pretraining dataset.
\newblock \emph{arXiv preprint arXiv:2412.02595}, 2024.

\bibitem[Team et~al.(2025)Team, Bai, Bao, Chen, Chen, Chen, Chen, Chen, Chen,
  Chen, et~al.]{team2025kimi}
Kimi Team, Yifan Bai, Yiping Bao, Guanduo Chen, Jiahao Chen, Ningxin Chen,
  Ruijue Chen, Yanru Chen, Yuankun Chen, Yutian Chen, et~al.
\newblock Kimi k2: Open agentic intelligence.
\newblock \emph{arXiv preprint arXiv:2507.20534}, 2025.

\bibitem[Team(2025)]{qwq32b}
Qwen Team.
\newblock Qwq-32b: Embracing the power of reinforcement learning, March 2025.
\newblock URL \url{https://qwenlm.github.io/blog/qwq-32b/}.

\bibitem[Touvron et~al.(2023{\natexlab{a}})Touvron, Lavril, Izacard, Martinet,
  Lachaux, Lacroix, Rozi{\`e}re, Goyal, Hambro, Azhar,
  et~al.]{touvron2023llama}
Hugo Touvron, Thibaut Lavril, Gautier Izacard, Xavier Martinet, Marie-Anne
  Lachaux, Timoth{\'e}e Lacroix, Baptiste Rozi{\`e}re, Naman Goyal, Eric
  Hambro, Faisal Azhar, et~al.
\newblock Llama: Open and efficient foundation language models.
\newblock \emph{arXiv preprint arXiv:2302.13971}, 2023{\natexlab{a}}.

\bibitem[Touvron et~al.(2023{\natexlab{b}})Touvron, Martin, Stone, Albert,
  Almahairi, Babaei, Bashlykov, Batra, Bhargava, Bhosale,
  et~al.]{touvron2023llama2}
Hugo Touvron, Louis Martin, Kevin Stone, Peter Albert, Amjad Almahairi, Yasmine
  Babaei, Nikolay Bashlykov, Soumya Batra, Prajjwal Bhargava, Shruti Bhosale,
  et~al.
\newblock Llama 2: Open foundation and fine-tuned chat models.
\newblock \emph{arXiv preprint arXiv:2307.09288}, 2023{\natexlab{b}}.

\bibitem[Wang et~al.(2022{\natexlab{a}})Wang, Wei, Schuurmans, Le, Chi, Narang,
  Chowdhery, and Zhou]{wang2022selfconsistency}
Xuezhi Wang, Jason Wei, Dale Schuurmans, Quoc Le, Ed~Chi, Sharan Narang,
  Aakanksha Chowdhery, and Denny Zhou.
\newblock Self-consistency improves chain of thought reasoning in language
  models.
\newblock \emph{arXiv preprint arXiv:2203.11171}, 2022{\natexlab{a}}.

\bibitem[Wang et~al.(2022{\natexlab{b}})Wang, Kordi, Mishra, Liu, Smith,
  Khashabi, and Hajishirzi]{wang2022self}
Yizhong Wang, Yeganeh Kordi, Swaroop Mishra, Alisa Liu, Noah~A Smith, Daniel
  Khashabi, and Hannaneh Hajishirzi.
\newblock Self-instruct: Aligning language models with self-generated
  instructions.
\newblock \emph{arXiv preprint arXiv:2212.10560}, 2022{\natexlab{b}}.

\bibitem[Wei et~al.(2022)Wei, Wang, Schuurmans, Bosma, Xia, Chi, Le, Zhou,
  et~al.]{wei2022chain}
Jason Wei, Xuezhi Wang, Dale Schuurmans, Maarten Bosma, Fei Xia, Ed~Chi, Quoc~V
  Le, Denny Zhou, et~al.
\newblock Chain-of-thought prompting elicits reasoning in large language
  models.
\newblock \emph{Advances in neural information processing systems},
  35:\penalty0 24824--24837, 2022.

\bibitem[Yang et~al.(2025)Yang, Li, Yang, Zhang, Hui, Zheng, Yu, Gao, Huang,
  Lv, et~al.]{yang2025qwen3}
An~Yang, Anfeng Li, Baosong Yang, Beichen Zhang, Binyuan Hui, Bo~Zheng, Bowen
  Yu, Chang Gao, Chengen Huang, Chenxu Lv, et~al.
\newblock Qwen3 technical report.
\newblock \emph{arXiv preprint arXiv:2505.09388}, 2025.

\bibitem[Yao et~al.(2023)Yao, Yu, Zhao, Shafran, Griffiths, Cao, and
  Narasimhan]{yao2023tree}
Shunyu Yao, Dian Yu, Jeffrey Zhao, Izhak Shafran, Tom Griffiths, Yuan Cao, and
  Karthik Narasimhan.
\newblock Tree of thoughts: Deliberate problem solving with large language
  models.
\newblock \emph{Advances in neural information processing systems},
  36:\penalty0 11809--11822, 2023.

\bibitem[Zhang et~al.(2025{\natexlab{a}})Zhang, Dong, Wu, Huang, Jia, Fernando,
  Shou, Zhang, and Liu]{zhang2025physreason}
Xinyu Zhang, Yuxuan Dong, Yanrui Wu, Jiaxing Huang, Chengyou Jia, Basura
  Fernando, Mike~Zheng Shou, Lingling Zhang, and Jun Liu.
\newblock Physreason: A comprehensive benchmark towards physics-based
  reasoning.
\newblock \emph{arXiv preprint arXiv:2502.12054}, 2025{\natexlab{a}}.

\bibitem[Zhang et~al.(2025{\natexlab{b}})Zhang, Ma, Gu, Yang, Zhuang, Wang,
  Huang, Wang, Huang, Song, et~al.]{zhang2025abench}
Yiming Zhang, Yingfan Ma, Yanmei Gu, Zhengkai Yang, Yihong Zhuang, Feng Wang,
  Zenan Huang, Yuanyuan Wang, Chao Huang, Bowen Song, et~al.
\newblock Abench-physics: Benchmarking physical reasoning in llms via
  high-difficulty and dynamic physics problems.
\newblock \emph{arXiv preprint arXiv:2507.04766}, 2025{\natexlab{b}}.

\bibitem[Zheng et~al.(2023)Zheng, Koh, Ju, Nguyen, May, Webb, and
  Pan]{zheng2023large}
Yizhen Zheng, Huan~Yee Koh, Jiaxin Ju, Anh~TN Nguyen, Lauren~T May, Geoffrey~I
  Webb, and Shirui Pan.
\newblock Large language models for scientific synthesis, inference and
  explanation.
\newblock \emph{arXiv preprint arXiv:2310.07984}, 2023.

\end{thebibliography}
\bibliographystyle{iclr2026_conference}

\appendix
\section{Appendix}
\subsection{Prompts}

We applied the LOCA workflow to a range of benchmark problems, employing a structured set of prompt templates to delegate distinct tasks to specialized agents. These tasks correspond to three primary roles: augmentation, review, and generation.

The prompt design for each role in the workflow is detailed below.

The proposed solution is first forwarded to the augmentation agent. Its responsibility is to enhance the solution by supplementing the reasoning chain and addressing any issues identified in the provided bug report. If no bug report is provided, the corresponding input field remains empty, and the agent proceeds without explicit error corrections.

\begin{tcblisting}{
    title=Augmentation Prompt,
    colback=white,
    colframe=green!50!black,
    fonttitle=\bfseries,
    breakable,
    listing only, 
    listing options={basicstyle=\ttfamily, breaklines=true} 
}
You are an AI expert specializing in physics problem-solving. Your task is to take a given physics problem, its potentially incomplete solution (Note that the original answer is not necessarily incomplete) and a report of bugs in this solution (Note that if the report shows no bugs, it is not necessarily no bugs), and produce a complete, step-by-step 'Refined Solution'. This refined solution MUST fix ALL bugs in the report, by adding or modifying missing steps, statements, principles, derivations, etc., to make the solution completed according to the problem statement and also make the logic clear, rigorous, and self-contained.
You should Not modify the final answer unless there is an inevitable contradiction during the derivation process.

You MUST adhere to the following strict formatting rules for your output:

1.  **Start Marker**: The entire output MUST begin *exactly* with the line `# Refined Solution` and nothing before it.
2.  **Sectioning**: The solution must be structured into sections using `###` headings.
3.  **First Section**: The first section MUST be titled `### Problem Statement Explanation`. In this section, you should detail and organize the information given in the problem statement, describe the physical situation / process, define all relevant variables one-by-one, and state any overall assumptions (e.g.,  approximations, geometric relations). You can not ignore any information in the problem statement.
4.  **Step Sections**: All following sections MUST be titled `### Step XXX`, where `XXX` is a number starting from 1 (e.g., `### Step 1`, `### Step 2`, ...). These sections should present the logical flow of the solution.
5.  **Content of Step Sections**: In the given solution, some parts are derivations or calculations, while others DIRECTLY adopt / introduce original formulas / assumptions / theorems / principles / geometric relations / boundary conditions. Within each `### Step XXX` section, you must follow this two-part structure:
-   **Principles/Original Formulas/Assumptions**: State all the original formulas / assumptions / theorems / physical principles / geometric relations / approximations / boundary conditions in their most original / general / universal / standard forms(i.e., according to physical theories / principles) being used in that step (maybe more than 1). Note that this must be in their most original, general, universal, or standard forms. Each of them MUST be placed on its own line and enclosed within a `$$\boxed{}$$` environment in a single line.
-   **Derivation**: Provide the subsequent mathematical derivation that APPLIES this original formulas, assumptions, physical principles, etc., to the problem(except for the review of the principle itself). This derivation MUST be enclosed within an `align` environment (e.g., `$$\begin{align} ... \end{align}$$`) with \label{} and \tag{} for reference.
6.  **Final Section**: The very last section of your output MUST be titled `### Final Answer`. In this section, you must clearly state the final result, and the mathematical expression for the answer MUST be enclosed within a `$$\begin{align} ... \end{align}$$` environment.

For example, given this problem and incomplete solution:
# Problem Statement
Find the acceleration 'a' of an apple with mass 'm', free falling near the ground with gravitational acceleration constant 'g'.
# Solution
The gravity on the apple is F_G = mg. Using Newton's second law F = ma, the acceleration is a = g.
Your refined solution should look like this:
# Refined Solution
### Problem Statement Explanation
This problem asks for the acceleration `a` of an apple of mass `m`. The apple is in a state of free fall near the Earth's surface, where the gravitational acceleration is a constant `g`. We assume that air resistance is negligible. The goal is to express `a` in terms of `m` and `g`.

### Step 1: Identify the Net Force on the Apple
First, we identify all forces acting on the apple. In a free-fall scenario where air resistance is neglected, the only force is the gravitational force exerted by the Earth. The formula for gravitational force near the Earth's surface is:
$$
\boxed{F_g = mg}
$$
Therefore, the net force acting on the apple is equal to the gravitational force.
$$
\begin{align}
F_{\text{net}} = F_g = mg
\label{eq:net_force} \tag{1}
\end{align}
$$

### Step 2: Apply Newton's Second Law of Motion
Next, we relate the net force on an object to its acceleration using Newton's Second Law of Motion. The standard form of the law is:
$$
\boxed{F_{\text{net}} = ma}
$$
We can now substitute the net force we found in Step 1 into this equation.
$$
\begin{align}
ma &= F_{\text{net}} \nonumber \ 
ma &= mg \quad (\text{using eq. \ref{eq:net_force}}) \nonumber \ 
a &= g
\end{align}
$$
Thus, the acceleration of the apple is equal to the gravitational acceleration constant `g`.

### Final Answer:
$$
\begin{align}
\boxed{a = g}
\end{align}
$$


Now I will give you the physics problem, its potentially incomplete solution and a report of bugs in this solution. Please fix ALL bugs to get a refine solution.

# Problem Statement
{question_statement}

# Solution
{solution}

# Bugs Report
{bugs_report}

 # Refined Solution
\end{tcblisting}

Subsequently, the review agent is tasked with identifying any issues in the augmented solution. 

\begin{tcblisting}{
    title=Reviewer Prompt,
    colback=white,
    colframe=green!50!black,
    fonttitle=\bfseries,
    breakable,
    listing only, 
    listing options={basicstyle=\ttfamily, breaklines=true} 
}
You are an AI expert specializing in reviewing advanced physics problems. Your task is to critically review the solution provided. You must review the solution step-by-step following the "### Step XXX" structure provided in the given solution. Carefully examine each step in the given solution. Your explanation of the error(s) must be concise, clear, and accurate.

Here is the problem and the solution to be reviewed:

# Problem Statement
{problem_statement}

# Solution
{solution}

# Review Instruction
When specifically reviewing each step in the given solution, please follow this instruction precisely:
---
{instruction}
---
After the step-by-step review, in the last line, write only 'Correct' if every step is correct, or 'Wrong' if you find any errors.
\end{tcblisting}

The reviewer receives distinct instructions to evaluate the solution from multiple perspectives—such as assumptions and derivations—ensuring a comprehensive assessment.

\begin{tcblisting}{
    title=Assumption Review Instruction,
    colback=white,
    colframe=orange!50!black,
    fonttitle=\bfseries,
    breakable,
    listing only,
    listing options={basicstyle=\ttfamily, breaklines=true}
}
In the given solution, some parts are derivations or calculations, while others DIRECTLY adopt / introduce original formulas / assumptions / theorems / principles / geometric relations / boundary conditions. Your task is to judge the original formulas / assumptions / theorems / principles / geometric relations / boundary conditions presented / stated in the target step.

1.  **Identification**:  Mainly focus on statements within the `\[\boxed{ }\]` environment along with other potential nature language statements in the provided step. 
2.  **Analyze its Validity**: Critically evaluate all the original formulas / assumptions / theorems / principles / geometric relations / boundary conditions based on the problem statement and the preceding (assumed correct) steps.
    - Is it a factually correct statement of an original formula / assumption / theorem / principle / geometric relation / boundary condition in its general form?
    - Is it an appropriate original formula / assumption / theorem / principle / geometric relation / boundary condition to apply at this specific stage of the problem?
.   - Is it specifically valid for this problem according to the problem statement?
   - Is it easily derived from more fundamental principles for this problem according to the problem statement, and therefore should not be used as an assumption?    - Is it specifically valid for this problem according to the problem statement?
3.  **Report Findings**:
    - If the original formula / assumption / theorem / principle / geometric relation / boundary condition is correctly stated and appropriately applied, state this clearly.
    - If any original formula / assumption / theorem / principle / geometric relation / boundary condition is misstated (e.g., a sign error in the original formula) or inappropriately applied (e.g., using a non-relativistic formula in a relativistic problem) or not suitable as an assumption, you MUST clearly and concisely describe the error.
4.  **Conclusion**: On the final line of your response, write **only** `Correct` if all the original formulas / assumptions / theorems / principles / geometric relations / boundary conditions are valid and correctly applied, or `Wrong` if there is any error.

**Example 1 (Correct Application):**
*Step to review contains:* `\[\boxed{F_{\text{net}} = ma}\]` *for a dynamics problem.*
*Your review:*
The principle stated in the boxed environment is `F_net = ma`, which is Newton's Second Law of Motion. This is a fundamental and correct principle for analyzing the dynamics of a massive object. Its application is appropriate for this step.
Correct

**Example 2 (Incorrect Application):**
*Problem involves a block on a 30-degree incline. Step to review contains:* `\[\boxed{N = mg}\]` *where N is the normal force.*
*Your review:*
The principle stated is `N = mg`. While this correctly relates normal force to gravitational force for an object on a horizontal surface, it is incorrectly applied here. In this specific problem, for a block on an inclined plane at an angle `\theta`, the correct relation for the normal force is `N = mg \cos\theta`. The assumption is therefore wrong for the given geometric conditions.
Wrong
\end{tcblisting}

\begin{tcblisting}{
    title=Derivation Review Instruction,
    colback=white,
    colframe=orange!50!black,
    fonttitle=\bfseries,
    breakable,
    listing only,
    listing options={basicstyle=\ttfamily, breaklines=true}
}
In the given solution, some parts are derivations or calculations, while others DIRECTLY adopt / introduce original formulas / assumptions / theorems / principles / geometric relations / boundary conditions. Your task is to judge the mathematical derivation / calculation within the target step.

1.  **Identify the Derivation**: Focus mainly on the equations inside the `\[\begin{ align} ... \end{align}\]` environment. You should not focus on the original formulas / assumptions / theorems / principles / geometric relations / boundary conditions stated within the `\[\boxed{ }\]` environment.
2.  **Analyze Step-by-Step**: Go through the derivation line by line. Verify each algebraic manipulation, substitution of variables from previous steps (as referenced by `\label{ }` or `\tag{ }`), and any numerical calculations.
3.  **Report Findings**:
    - If the entire derivation is mathematically sound, state this.
    - If you find an error, you MUST pinpoint the exact line or transition where the error occurs. Quote the incorrect part, explain the mistake (e.g., "algebraic error," "incorrect substitution from tag{3}," "a sign was dropped", etc.) and state what the correct derivation or result should be. Use the `label` or `tag` for reference if available.
4.  **Conclusion**: On the final line of your response, write **only** `Correct` if the derivation is flawless, or `Wrong` if you find any mathematical error.

**Example 1 (Correct Derivation):**
*Step to review contains:* `\[\begin{align} ma &= F_{\text{net}} \nonumber \ ma &= mg \quad (\text{using eq. \ref{eq:net_force}}) \nonumber \ a &= g \end{align}\]`
*Your review:*
The derivation begins by correctly substituting `F_net` with `mg` based on the reference to `eq:net_force`. The final step correctly isolates `a` by dividing both sides of `ma = mg` by `m`. The derivation is mathematically sound.
Correct

**Example 2 (Incorrect Derivation):**
*Step to review contains:* `\[\begin{align} \frac{1}{2}mv_f^2 - \frac{1}{2}mv_i^2 &= mgh \label{eq:energy} \tag{3} \ v_f^2 - v_i^2 &= \frac{1}{2}gh \end{align}\]`
*Your review:*
The derivation starts from the work-energy theorem, referenced as tag{3}. In the transition from the first line to the second, the term `m` is canceled from the left side, but the right side is incorrectly divided by `2m` instead of just `m`. The correct second line should be `v_f^2 - v_i^2 = 2gh`. This is an algebraic error.
Wrong    
\end{tcblisting}

The outputs from these reviews are integrated into a standardized issues report template.

\begin{tcblisting}{
    title=Issues Report Template,
    colback=white,
    colframe=orange!50!black,
    fonttitle=\bfseries,
    breakable,
    listing only, 
    listing options={basicstyle=\ttfamily, breaklines=true} 
}
# Issues found in solution
## judge assumption
{issues_about_assumption}
## judge derivation
{issues_about_derivation}
\end{tcblisting}

Finally, the completed issues report is forwarded to the secretary agent for summarization, whose role is to consolidate and finalize the report.

\begin{tcblisting}{
    title=Summarization Prompt,
    colback=white,
    colframe=green!50!black,
    fonttitle=\bfseries,
    breakable,
    listing only,
    listing options={basicstyle=\ttfamily, breaklines=true}
}
You are an AI expert specializing in advanced physics questions. Please extract, in a concise, clear and accurate manner, errors from a given review of a certain physics question. The errors, which should be extracted, have been recorded, and you MUST list them one by one in the format of * * Incorrect Part: * * and * * Explanation of Mistake: * *, in order to provide a clear error report.
The review text are as follows: {review}.
\end{tcblisting}
The final report can be fed back to the augmentation agent as constructive feedback, forming an iterative review loop that enables continuous refinement of the solution through multiple cycles of augmentation and critical evaluation.

\newpage

In the ablation experiment, the review prompt and augmentation prompt have slight differences, as shown below.

\begin{tcblisting}{
    title=Augmentation Prompt,
    colback=white,
    colframe=gray!50!black,
    fonttitle=\bfseries,
    breakable,
    listing only,
    listing options={basicstyle=\ttfamily, breaklines=true}
}
You are a physics expert tasked with improving a solution to a physics problem based on review feedback.

**Problem:**
{question_statement}

**Current Solution:**
{solution}

**Review Feedback:**
{feedback}

Please provide an improved solution that addresses the issues mentioned in the feedback. The very last section of your output MUST be titled `### Final Answer`. Provide the final answer at the end in Latex boxed format $$\boxed{}$$. Example: $$ \boxed{ final_answer} $$
\end{tcblisting}

\begin{tcblisting}{
    title=Reviewer Prompt,
    colback=white,
    colframe=gray!50!black,
    fonttitle=\bfseries,
    breakable,
    listing only,
    listing options={basicstyle=\ttfamily, breaklines=true}
}
You are an AI expert specializing in physics problem solving. Your task is to determine whether the final answer provided in a physics solution is correct.

You will be given:
1. A physics problem statement
2. A complete solution to that problem

Your job is to evaluate whether the final answer is mathematically and physically correct based on the problem requirements and the solution provided.

Here is the information to review:

# Problem Statement
{question_statement}

# Complete Solution
{solution}

# Instructions
Provide your analysis, then in the last line, MUST write only 'Correct' if the final answer is correct, or 'Wrong' if the final answer is incorrect.
\end{tcblisting}

\subsection{Examples of logical chain augmentation}
This typical example is based on question-250 from PHYbench, which is a Medium difficulty question of electromagnetic. The question and raw answer are following:

\begin{mytextbox}[title=Question]
There is now an electrolyte with thickness $L$ in the $z$ direction, infinite in the $x$ direction, and infinite in the $y$ direction. The region where $y > 0$ is electrolyte 1, and the region where $y < 0$ is electrolyte 2. The conductivities of the two dielectrics are $\sigma_{1}, \sigma_{2}$, and the dielectric constants are $\varepsilon_{1}, \varepsilon_{2}$, respectively. On the $xOz$ interface of the two dielectrics, two cylindrical holes with a radius $R$ are drilled in the $z$ direction, spaced $2D\ (D > R, R, D \ll L)$ apart, with centers located on the interface as long straight cylindrical holes. Two cylindrical bodies $\pm$ are inserted into the holes, with the type of the cylinders given by the problem text below.

The cylindrical bodies $\pm$ are metal electrodes filling the entire cylinder. Initially, the system is uncharged, and at $t=0$, a power source with an electromotive force $U$ and internal resistance $r_{0}$ is used to connect the electrodes. Find the relationship between the current through the power source and time, denoted as $i(t)$.

\end{mytextbox}

\begin{mytextbox}[title=Raw Answer]
Given the potential difference $u$, it can be seen:
$$
\varphi_{+}=u/2,\varphi_{-}=-u/2,\lambda=\frac{2\pi(\varepsilon_{1}+\varepsilon_{2})\varphi_{+}}{2\xi_{+}}=\frac{\pi(\varepsilon_{1}+\varepsilon_{2})u}{\operatorname{arccosh}(\mathrm{D/R})}
$$  
Select a surface encapsulating the cylindrical surface and examine Gauss's theorem. For the positive electrode, it is easy to see:
$$
\iint\vec{E}\cdot d\vec{S}=L\oint\vec{E}\cdot\hat{n}d l=\frac{\lambda L}{(\varepsilon_{1}+\varepsilon_{2})/2}=\frac{2\pi u L}{2\mathrm{arccosh}(D/R)}
$$  
Since the above potential distribution is deemed directly applicable for the calculation of current, the total current flowing out of the positive electrode is:
$$
I=\iint\sigma\vec{E}\cdot d\vec{S}=\frac{\sigma_{1}+\sigma_{2}}{2}\times\frac{2\pi u L}{2\mathrm{arccosh}(D/R)}
$$  
Given the current $i$ passing through the power source, this current changes the net charge:
$$
{\frac{d(\lambda L)}{d t}}=i-I=i-{\frac{2\pi u L}{2\mathrm{arccosh}(D/R)}}={\frac{\pi(\varepsilon_{1}+\varepsilon_{2})L}{2\mathrm{arccosh}(D/R)}}{\frac{d u}{d t}}
$$
According to the loop voltage drop equation:
\begin{align*}
    &U=r_{0}i+u\rightarrow u=U-r_{0}i\\
    \rightarrow & i-\frac{\pi(\sigma_{1}+\sigma_{2})L}{2\mathrm{arccosh}(D/R)}(U-r_{0}i) = -\frac{\pi(\varepsilon_{1}+\varepsilon_{2})L}{2\mathrm{arccosh}(D/R)} r_{0} \frac{d i}{d t}\\
    \rightarrow & \frac{d i}{d t} = \frac{(\sigma_{1}+\sigma_{2})U}{r_{0}(\varepsilon_{1}+\varepsilon_{2})}-\left(\frac{\sigma_{1}+\sigma_{2}}{\varepsilon_{1}+\varepsilon_{2}}+\frac{2\mathrm{arccosh}(D/R)}{\pi r_{0}L(\varepsilon_{1}+\varepsilon_{2})}\right)i
\end{align*}

At time $t=0$, all current should preferentially enter the capacitor. At this time, the initial current is $U/r_{0}$, and this differential equation yields:
$$
i(t)=\frac{U}{r_{0}\left(1+\frac{2\mathrm{arccosh}(D/R)}{\pi r_{0}L(\sigma_{1}+\sigma_{2})}\right)}\left\{2+\frac{2\mathrm{arccosh}(D/R)}{\pi r_{0}L(\sigma_{1}+\sigma_{2})}-\exp{\left[-\left(\frac{\sigma_{1}+\sigma_{2}}{\varepsilon_{1}+\varepsilon_{2}}+\frac{2\mathrm{arccosh}(D/R)}{\pi r_{0}L(\varepsilon_{1}+\sigma_{2})}\right)\right]}\right\},
$$
\end{mytextbox}

One can see missing logical steps like definitions of $u$, $\xi_{+}$, $\lambda$ and too brief mathematical derivations. At the first loop, LOCA's review agent gives the bug report as:

\begin{mytextbox}[title=Bug Report 1]
Here is a report of the errors extracted from the provided review:

*   **Incorrect Part:**

The foundational formula for linear charge density, $\lambda=\frac{\pi(\varepsilon_{1}+\varepsilon_{2})u}{\operatorname{arccosh}(\mathrm{D/R})}$.

*   **Explanation of Mistake:**

This formula is physically incorrect for the described system. The correct formula for capacitance per unit length, from which $\lambda$ is derived, should have $\operatorname{arccosh}(D/(2R))$ in the denominator, not $\operatorname{arccosh}(D/R)$. Furthermore, the correct formula has an additional factor of $2$ in the denominator.

*   **Incorrect Part:**

The algebraic derivation of the linear charge density $\lambda$: $$\frac{2\pi(\varepsilon_{1}+\varepsilon_{2})\varphi_{+}}{2\xi_{+}}=\frac{\pi(\varepsilon_{1}+\varepsilon_{2})u}{\operatorname{arccosh}(\mathrm{D/R})}.$$

*   **Explanation of Mistake:**

There is an algebraic error in this step. Substituting $\varphi_+ = u/2$ and $\xi_+ = \operatorname{arccosh}(D/R)$ into the middle expression yields $\frac{\pi(\varepsilon_{1}+\varepsilon_{2})u}{2\operatorname{arccosh}(\mathrm{D/R})}$, which differs from the expression on the right by a factor of 2 in the denominator.

*   **Incorrect Part:**

The calculation of the electric flux integral: $$\frac{\lambda L}{(\varepsilon_{1}+\varepsilon_{2})/2}=\frac{2\pi u L}{2\mathrm{arccosh}(D/R)}.$$

*   **Explanation of Mistake:**

An algebraic error was made. Substituting the solution's (already incorrect) formula for $\lambda$ into the left side of the equation results in $\frac{2\pi u L}{\operatorname{arccosh}(\mathrm{D/R})}$. The expression on the right side of the equality has an erroneous extra factor of 2 in the denominator.

*   **Incorrect Part:**

The expression for the leakage current $I$ used in the charge conservation equation: $$i-I=i-{\frac{2\pi u L}{2\mathrm{arccosh}(D/R)}}.$$

*   **Explanation of Mistake:**

This is an incorrect substitution. The expression used for $I$ is inconsistent with the one derived in the previous step $$I=\frac{\sigma_{1}+\sigma_{2}}{2}\times\frac{2\pi u L}{2\mathrm{arccosh}(D/R)}.$$
It is physically incorrect as it omits the crucial effective conductivity term $(\sigma_1+\sigma_2)/2$.

*   **Incorrect Part:**

The final provided solution for the current $i(t)$.

*   **Explanation of Mistake:**

The final expression for $i(t)$ is mathematically incorrect because it does not satisfy the physical conditions of the circuit. Specifically, its steady-state value (as $t \to \infty$) does not match the correct steady-state current, which should be $i_{ss} = U/(r_0+R_{elec})$.
\end{mytextbox}

Due to the excessive logical leaps in the raw answer, the review agent suggests that there are multiple incorrect parts, such as definition of linear charge density $\lambda$, the algebraic derivations and final solution. 

After augmentation, the logical leaps are fixed to providing a refined answer:
\begin{mytextbox}[title=Refined Answer 1]

Problem Statement Explanation

The problem describes a physical system composed of two semi-infinite electrolytes, separated by the $xOz$ plane.

-   **Electrolyte 1**: Occupies the region $y > 0$, with electrical conductivity $\sigma_1$ and dielectric constant (permittivity) $\varepsilon_1$.

-   **Electrolyte 2**: Occupies the region $y < 0$, with conductivity $\sigma_2$ and permittivity $\varepsilon_2$.

Two long, parallel cylindrical metal electrodes, denoted as $+$ and $-$, are placed within this system.

-   **Geometry**: The electrodes are of radius $R$ and length $L$. They are oriented parallel to the $z$-axis. Their centers lie on the $xOz$ interface and are separated by a distance of $2D$.

-   **Assumptions**: The geometry satisfies the conditions $D > R$ and $\{R, D\} \ll L$. The latter condition allows us to neglect end effects and treat the problem as two-dimensional in the $xy$-plane, with total quantities (like capacitance and resistance) being proportional to the length $L$.

At time $t=0$, this system is connected to a power source.

-   **Power Source**: An ideal electromotive force (EMF) $U$ with a constant internal resistance $r_0$.

-   **Initial Condition**: The system is initially uncharged, meaning the potential difference across the electrodes is zero at $t=0$.

The goal is to find the current $i(t)$ flowing from the power source as a function of time.

Step 1: Equivalent Circuit Model

The physical system can be modeled as a simple electrical circuit. The two electrodes immersed in the electrolytes act as a capacitor, storing charge, and simultaneously as a resistor, allowing a leakage current to flow between them through the conductive medium.

-   The capacitive nature is due to the storage of electric charge on the electrodes when a potential difference is applied, with the electrolytes acting as the dielectric material. Let's denote the equivalent capacitance as $C$.

-   The resistive nature is due to the flow of charge (ions) through the electrolytes under the influence of the electric field, from the positive to the negative electrode. Let's denote the equivalent resistance of the electrolytes as $R_{elec}$.

Since charge can be stored on the electrodes while simultaneously leaking through the medium, the capacitor $C$ and the resistor $R_{elec}$ are in parallel. This parallel combination is connected in series with the internal resistance $r_0$ of the power source with EMF $U$.

Step 2: Derivation of the System's Capacitance

The total capacitance is the sum of the capacitances of the upper and lower halves of the system, which are connected in parallel.

First, we state the standard formula for the capacitance per unit length, $C'$, between two parallel cylindrical conductors of radius $R$ with centers separated by a distance $2D$ in a uniform dielectric medium with permittivity $\varepsilon$.
$$
\boxed{C' = \frac{\pi\varepsilon}{\operatorname{arccosh}(D/R)}}
$$
In our problem, the upper half-space ($y>0$) is filled with electrolyte 1 ($\varepsilon_1$), and the lower half-space ($y<0$) is filled with electrolyte 2 ($\varepsilon_2$). Due to the symmetry of the electric field about the $y=0$ plane, we can consider this as two capacitors in parallel.

-   $C_1$: Capacitance of the upper half, in a medium with permittivity $\varepsilon_1$.

-   $C_2$: Capacitance of the lower half, in a medium with permittivity $\varepsilon_2$.

The capacitance per unit length for each half is:
\begin{align}
C'_1 = \frac{\pi\varepsilon_1}{\operatorname{arccosh}(D/R)} \label{eq:C1_prime} \tag{1} \\
C'_2 = \frac{\pi\varepsilon_2}{\operatorname{arccosh}(D/R)} \label{eq:C2_prime} \tag{2}
\end{align}

The total capacitance per unit length, $C'_{total}$, is the sum of the parallel contributions.
$$
\boxed{C'_{total} = C'_1 + C'_2}
$$
The total capacitance $C$ for the electrode length $L$ is then derived.
\begin{align}
C'_{total} &= \frac{\pi\varepsilon_1}{\operatorname{arccosh}(D/R)} + \frac{\pi\varepsilon_2}{\operatorname{arccosh}(D/R)} = \frac{\pi(\varepsilon_1 + \varepsilon_2)}{\operatorname{arccosh}(D/R)} \label{eq:C_prime_total} \tag{3} \\
C &= C'_{total} \cdot L = \frac{\pi(\varepsilon_1 + \varepsilon_2)L}{\operatorname{arccosh}(D/R)} \label{eq:C_total} \tag{4}
\end{align}

Step 3: Derivation of the System's Resistance
Similarly, the total resistance of the electrolyte is determined by the two parallel paths for current flow through the upper and lower electrolytes.

For a system with a given geometry, the resistance $R$ and capacitance $C$ in a homogeneous medium are related.
$$
\boxed{R C = \frac{\varepsilon}{\sigma}}
$$
We can apply this relation to each half of our system.
-   $R_1$: Resistance of the upper electrolyte with conductivity $\sigma_1$.
-   $R_2$: Resistance of the lower electrolyte with conductivity $\sigma_2$.
\begin{align}
R_1 &= \frac{\varepsilon_1}{\sigma_1 C_1} = \frac{\varepsilon_1}{\sigma_1 (C'_1 L)} = \frac{\varepsilon_1}{\sigma_1 L} \frac{\operatorname{arccosh}(D/R)}{\pi\varepsilon_1} = \frac{\operatorname{arccosh}(D/R)}{\pi\sigma_1 L} \label{eq:R1} \tag{5} \\
R_2 &= \frac{\varepsilon_2}{\sigma_2 C_2} = \frac{\varepsilon_2}{\sigma_2 (C'_2 L)} = \frac{\varepsilon_2}{\sigma_2 L} \frac{\operatorname{arccosh}(D/R)}{\pi\varepsilon_2} = \frac{\operatorname{arccosh}(D/R)}{\pi\sigma_2 L} \label{eq:R2} \tag{6}
\end{align}
Since the current can flow through both electrolytes simultaneously, these two resistances are in parallel. The total equivalent resistance of the electrolyte, $R_{elec}$, is given by:
$$
\boxed{\frac{1}{R_{elec}} = \frac{1}{R_1} + \frac{1}{R_2}}
$$
\begin{align}
\frac{1}{R_{elec}} &= \frac{\pi\sigma_1 L}{\operatorname{arccosh}(D/R)} + \frac{\pi\sigma_2 L}{\operatorname{arccosh}(D/R)} = \frac{\pi(\sigma_1 + \sigma_2)L}{\operatorname{arccosh}(D/R)} \label{eq:inv_Relec} \tag{7} \\
R_{elec} &= \frac{\operatorname{arccosh}(D/R)}{\pi(\sigma_1 + \sigma_2)L} \label{eq:Relec} \tag{8}
\end{align}

Step 4: Formulation of the Governing Differential Equation

Let $i(t)$ be the current from the source and $u(t)$ be the potential difference across the electrodes. Applying Kirchhoff's Voltage Law to the circuit loop:
$$
\boxed{U = i(t)r_0 + u(t)}
$$
The current $i(t)$ from the source splits into two paths in the parallel combination: a charging current $i_C(t)$ for the capacitor and a leakage current $i_R(t)$ through the resistor.
$$
\boxed{i(t) = i_C(t) + i_R(t)}
$$
The currents $i_C(t)$ and $i_R(t)$ are defined by the properties of the capacitor and resistor:
$$
\boxed{i_C(t) = C \frac{du}{dt}}
$$
$$
\boxed{i_R(t) = \frac{u(t)}{R_{elec}}}
$$
Combining these, we get a differential equation relating $i(t)$ and $u(t)$.
\begin{align}
i(t) = C \frac{du}{dt} + \frac{u(t)}{R_{elec}} \label{eq:current_split} \tag{9}
\end{align}
To find an equation solely for $i(t)$, we eliminate $u(t)$. From the loop law, $u(t) = U - i(t)r_0$. Differentiating with respect to time gives $\frac{du}{dt} = -r_0 \frac{di}{dt}$. Substituting these into Eq. \eqref{eq:current_split}:
\begin{align}
i &= C \left(-r_0 \frac{di}{dt}\right) + \frac{U - i r_0}{R_{elec}} \nonumber \\
i &= -C r_0 \frac{di}{dt} + \frac{U}{R_{elec}} - \frac{r_0}{R_{elec}} i \nonumber \\
C r_0 \frac{di}{dt} &= \frac{U}{R_{elec}} - i \left(1 + \frac{r_0}{R_{elec}}\right) \nonumber \\
\frac{di}{dt} &= \frac{U}{C r_0 R_{elec}} - \left(\frac{1}{C r_0} + \frac{1}{C R_{elec}}\right) i \label{eq:DE_for_i} \tag{10}
\end{align}
This is a first-order linear ordinary differential equation for $i(t)$.

Step 5: Solving the Differential Equation
The differential equation \eqref{eq:DE_for_i} is of the form $\frac{di}{dt} + B i = A$.
$$
\boxed{\frac{dy}{dt} + P(t)y = Q(t) \implies y(t) = e^{-\int P(t)dt} \left( \int Q(t) e^{\int P(t)dt} dt + K \right)}
$$
For our constant-coefficient case, the general solution is $i(t) = i_{ss} + i_h(t) = A/B + K e^{-Bt}$, where $i_{ss}$ is the steady-state current and $K$ is a constant determined by the initial condition.
\begin{align}
B &= \frac{1}{C r_0} + \frac{1}{C R_{elec}} = \frac{r_0 + R_{elec}}{C r_0 R_{elec}} \label{eq:B_coeff} \tag{11} \\
i_{ss} &= \frac{A}{B} = \frac{U/(C r_0 R_{elec})}{(r_0 + R_{elec})/(C r_0 R_{elec})} = \frac{U}{r_0 + R_{elec}} \label{eq:iss} \tag{12}
\end{align}
The initial condition at $t=0$ is that the system is uncharged, so $u(0)=0$. From the loop law, $U = i(0)r_0 + u(0)$, which gives the initial current:
\begin{align}
i(0) = \frac{U}{r_0} \label{eq:i0} \tag{13}
\end{align}
Applying this to the general solution at $t=0$:
\begin{align}
i(0) &= i_{ss} + K e^0 \nonumber \\
\frac{U}{r_0} &= \frac{U}{r_0 + R_{elec}} + K \nonumber \\
K &= \frac{U}{r_0} - \frac{U}{r_0 + R_{elec}} = U \frac{(r_0 + R_{elec}) - r_0}{r_0(r_0 + R_{elec})} = \frac{U R_{elec}}{r_0(r_0 + R_{elec})} \label{eq:K_const} \tag{14}
\end{align}
Substituting $i_{ss}$, $K$, and $B$ back into the solution gives $i(t)$:
\begin{align}
i(t) = \frac{U}{r_0 + R_{elec}} + \frac{U R_{elec}}{r_0(r_0 + R_{elec})} \exp\left(- \frac{r_0 + R_{elec}}{C r_0 R_{elec}} t\right) \label{eq:i_t_general} \tag{15}
\end{align}

Final Answer

The relationship between the current through the power source and time, $i(t)$, is found by substituting the expressions for the equivalent capacitance $C$ (Eq. \eqref{eq:C_total}) and resistance $R_{elec}$ (Eq. \eqref{eq:Relec}) into the general solution (Eq. \eqref{eq:i_t_general}).

The steady-state current is:
$$
i_{ss} = \frac{U}{r_0 + R_{elec}} = \frac{U}{r_0 + \frac{\operatorname{arccosh}(D/R)}{\pi(\sigma_1 + \sigma_2)L}}
$$
The decay constant in the exponent is:
$$
B = \frac{1}{C R_{elec}} + \frac{1}{C r_0} = \frac{\sigma_1 + \sigma_2}{\varepsilon_1 + \varepsilon_2} + \frac{\operatorname{arccosh}(D/R)}{\pi r_0 (\varepsilon_1 + \varepsilon_2)L}
$$
The final expression for the current $i(t)$ is:
\begin{align}
\boxed{i(t) = \frac{U}{r_0 + \frac{\operatorname{arccosh}(D/R)}{\pi(\sigma_1 + \sigma_2)L}} + \left(\frac{U}{r_0} - \frac{U}{r_0 + \frac{\operatorname{arccosh}(D/R)}{\pi(\sigma_1 + \sigma_2)L}}\right) \exp\left[ - \left(\frac{\sigma_1 + \sigma_2}{\varepsilon_1 + \varepsilon_2} + \frac{\operatorname{arccosh}(D/R)}{\pi r_0 (\varepsilon_1 + \varepsilon_2)L}\right) t \right]}
\end{align}
\end{mytextbox}

This answer is obviously more detailed and easier to verify.
The review agent accurately locate the error:
\begin{mytextbox}[title=Bug Report 2]
Here is the error report based on the provided review:

* * Incorrect Part: * *

Step 2: Derivation of the System's Capacitance.

* * Explanation of Mistake: * *

The solution incorrectly applies the formula for capacitance per unit length between two full parallel cylinders,
$$C' = \frac{\pi\varepsilon}{\operatorname{arccosh}(D/R)},$$ to a system involving two half-spaces. The correct capacitance for each half-space is only half of this value. By omitting the factor of 1/2, the solution calculates a total capacitance `C` that is twice the correct value.
\end{mytextbox}

Referring to this bug report, LLM generated the correct answer with correct steps in the next iteration. This answer received recognition from both the reviewer agent and human expert. The correct answer is
\begin{mytextbox}[title=Correct Answer]
\begin{align*}
    i(t) &= \frac{U}{r_0 + \frac{2\operatorname{arccosh}(D/R)}{\pi(\sigma_1 + \sigma_2)L}}\\
    &+ \left(\frac{U}{r_0} - \frac{U}{r_0 + \frac{2\operatorname{arccosh}(D/R)}{\pi(\sigma_1 + \sigma_2)L}}\right) \exp\left[ - \left(\frac{\sigma_1 + \sigma_2}{\varepsilon_1 + \varepsilon_2} + \frac{2\operatorname{arccosh}(D/R)}{\pi r_0 (\varepsilon_1 + \varepsilon_2)L}\right) t \right],
\end{align*}
\end{mytextbox}
which is different from the raw answer. This example demonstrates that Logical Chain Augmentation not only improves the detection of errors within detailed solution steps, but also enables the LLM to generate the correct reasoning path and final answer.

\subsection{The Use of LLMs}
LLMs were used to help polish writing, improve conciseness, and check grammar across the sections of the paper.

\end{document}